\pdfoutput=1
\documentclass[11pt]{article}

\usepackage{verbatimbox}
\usepackage[preprint]{acl}
\usepackage{times}

\usepackage{latexsym}
\usepackage{geometry}
\usepackage{subcaption}
\usepackage[T1]{fontenc}
\usepackage[utf8]{inputenc}

\usepackage{microtype}

\usepackage{inconsolata}

\usepackage{graphicx}
\usepackage{longtable}
\usepackage{amsmath}

\title{Training Agents with Weakly Supervised Feedback from\\ Large Language Models}

\author{Dihong Gong\thanks{Authors with equally significant contribution.}, \ 
        Pu Lu\footnotemark[1], \
        Zelong Wang\footnotemark[1], \ 
        Meng Zhou, \ 
        Xiuqiang He \\
Tencent Inc, China\\
{\tt\small gongdihong@gmail.com, luput@foxmail.com, \{zelongwang, lincyzhou, xiuqianghe\}@tencent.com} 
}

\begin{document}
\maketitle
\begin{abstract}
Large Language Models (LLMs) offer a promising basis for creating agents that can tackle complex tasks through iterative environmental interaction. Existing methods either require these agents to mimic expert-provided trajectories or rely on definitive environmental feedback for reinforcement learning which limits their application to specific scenarios like gaming or code generation. This paper introduces a novel training method for LLM-based agents using weakly supervised signals from a critic LLM, bypassing the need for expert trajectories or definitive feedback. Our agents are trained in iterative manner, where they initially generate trajectories through environmental interaction. Subsequently, a critic LLM selects a subset of good trajectories, which are then used to update the agents, enabling them to generate improved trajectories in the next iteration. Extensive tests on the API-bank dataset show consistent improvement in our agents' capabilities and comparable performance to GPT-4, despite using open-source models with fewer parameters.
\end{abstract}

\section{Introduction}

The AI community has long pursued the goal of creating agents that can perform a broad range of tasks across various environments at a level comparable to humans, and substantial efforts have been invested in this direction \cite{nakano2021webgpt,chen2022program,gur2023real,xi2023rise,schick2024toolformer,shen2024hugginggpt}. Similar to human learning, an agent begins by learning basic knowledge and skills through imitation\cite{tang2023toolalpaca, zeng2023agenttuning,patil2023gorilla}. As the agent advances, it is anticipated to continually learn and adapt to new tasks by interacting with diverse environments\cite{zhang2024agent}.

Large language models (LLMs) are viewed as a promising basis for building such versatile agents due to their generalization capabilities, and numerous initiatives have been undertaken in this area. One approach primarily depends on human supervision, where LLM-based agents emulate expert-provided trajectories from various environments \cite{chen2023fireact,ToRA, Chen2024AgentFLANDD, zeng2023agenttuning, xu2023lemur}. While effective, this method relies on skilled annotators and substantial financial resources, making scalability a challenge.

Another research direction enables LLM-based agents to self-improve through environmental feedback, thereby reducing the reliance on human supervision \cite{wang2023math,huang2022large,kadlvcik2024self,shao2024deepseekmath} . This approach is rooted in the concept of reinforcement learning\cite{sutton2018reinforcement}, where agents learn to make decisions by experiencing the consequences of their actions. In this context, environmental feedback serves as a form of reward or penalty, guiding the agent towards optimal behavior. For instance, in coding tasks\cite{luo2023wizardcoder, jiang2023selfevolve,ni2023lever}, the agent might receive positive feedback when it writes a piece of code without errors, and negative feedback when the code fails to compile. Similarly, in gaming  scenarios\cite{schrittwieser2020mastering,berner2019dota,vinyals2019grandmaster}, the agent could be rewarded for winning a game or achieving a high score, and penalized for losing or performing poorly. However, a significant limitation of this approach is that it typically requires decisive environmental feedback. As a result, this method is often confined to narrow, well-defined tasks such as coding, gaming, or mathematical calculations\cite{huang2022large,wang2023math,kadlvcik2024self,liao2024mario}, where the results of the agent's actions can be unequivocally determined. This restricts the applicability of such agents, as many real-world tasks involve ambiguous outcomes and complex, multi-dimensional feedback.

In this study, we present a novel framework that enables agents to self-evolve iteratively, eliminating dependence on expert trajectories or decisive environmental feedback. Our agents initially interact with the environment, generating trajectories. A critic Large Language Model (LLM) evaluates these, selecting effective ones. Unlike decisive feedback, our critic LLM provides weak feedback applicable to a broad spectrum of applications. To mitigate errors from weak feedback, we use iterative training, selecting a few high-confidence successful trajectories. This enables comprehensive exploration and incremental learning from the critic LLM's feedback. Our primary contributions include:

1. Introduce a novel framework for the iterative training of agents under weak supervision via a critic LLM.

2. Demonstrating that agents can effectively evolve progressively using our training framework.

3. Achieving performance comparable to GPT-4 on the public benchmark dataset API-Bank \cite{li2023api}using significantly smaller LLMs.

\begin{figure*}[t]
  \includegraphics[width=\linewidth]{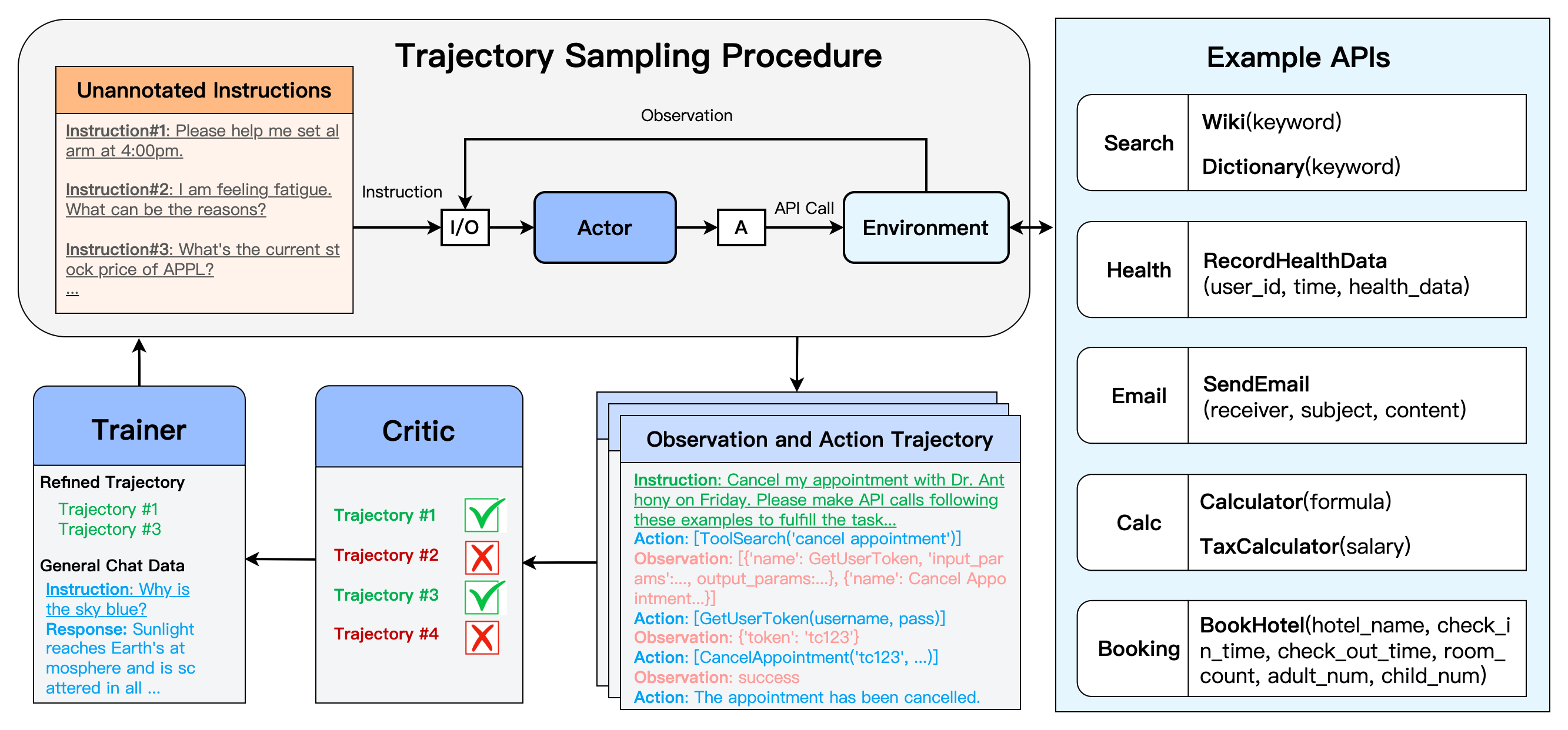}
  \caption{Our self-evolving algorithm employs a comprehensive training pipeline to instruct LLMs in the utilization of APIs. The process begins with a set of instructions, which guide the actor module in interacting with an environment composed of various APIs, thereby generating a sequence of trials. Subsequently, the critic module is applied to discern a subset of trials where it perceives the actor has successfully executed the instruction. These successful trials are then forwarded to the trainer module, which updates the underlying actor module. To prevent overfitting, this update is supplemented with general chat data. This procedure is iteratively repeated, allowing the actor module to evolve and adapt to its environment.}
  \label{fig:pipeline}
\end{figure*}

\section{Method}
%In this section, we delve into the specifics of our methodology. We begin by providing an overview of the entire process, outlining the general pipeline of our method. Following this, we dissect the crucial modules within this pipeline, including trajectory sampling, the critic module, and supervised fine-tuning, in the subsequent subsections.

\subsection{Overview}
Our agents adapt to their environment through a series of iterative learning procedures. As depicted in Figure~\ref{fig:pipeline}, given a set of instructions, the agents initially explore the environment involving multi-turn interactions with the environment, resulting in a set of trajectories. These trajectories are evaluated by the critic module. Higher scores are allocated to better trajectories. The scored trajectories are subsequently fed into the trainer module. This procedure is repeated over several rounds, allowing the agents to adapt to the environment progressively.

\subsection{Trajectory Sampling}
The trajectory sampling procedure empowers the agent to explore the environment comprehensively. The agent, also referred to as the 'actor' in Figure~\ref{fig:pipeline}, generates $K$ trajectories for each instruction. Consequently, for a set of $N$ instructions, , this procedure yields a total of $N \times K$ trajectories.

\begin{equation*}
\begin{split}
T = \{I^{(n,k)}A_{1}^{(n,k)}O_{1}^{(n,k)}\dots A_{M}^{(n,k)}O_{M}^{(n,k)} | \\
n=1 \dots N, k=1 \dots K\}
\end{split}
\label{eq:trajectory}
\end{equation*}

In the above equation, $I^{(n,k)}$ represents the $n-th$ instruction used for generating its corresponding $k-th$ trajectory. The symbol $A$ represents the action (e.g. API call) generated by agent, while $O$ denotes the observation (e.g. result of the API call). The maximum number of interaction steps is capped at $M$. A simplified trajectory is depicted in Figure~\ref{fig:pipeline}. For a detailed example of a full trajectory, including the prompt construction, please refer to the Appendix. This trajectory sampling procedure is a crucial part of our methodology, enabling the agent to learn from a wide range of experiences and interactions within the environment.

\subsection{Critic Module}

 The primary objective of this step is to identify a subset of trajectories that can be instrumental in enhancing the future performance of the agents. Each trajectory typically involves a multi-round interaction with the environment, where the agent responds to a given instruction through a series of API calls. Errors can occur at various stages along the trajectory, including API selection, API calling parameters, exception handling, and conclusion. This complexity makes it challenging to accurately assess the quality of individual trajectories.

To address this challenge, we have incorporated two key design considerations. Firstly, we utilize a Large Language Model (LLM) as the foundation for the critic module and design sophisticated prompts. This approach allows the critic module to evaluate the trajectories in a comprehensive manner. Secondly, we adopt a progressively iterative learning approach from environmental interactions. This strategy minimizes the risk of learning from poor-quality trajectories provided by the critic module. The detailed prompt for the critic is included in the Appendix section of this paper.

\subsection{Supervised Fine-Tuning}
In this stage, the agents adjust their model parameters in response to the training dataset. This adaptation process allows the agents to refine their performance, leading to the generation of higher-quality trajectories in subsequent rounds.

\textbf{Training Data.} During each round, we generate a collection of trajectories through interaction with the environment, which are then evaluated in the subsequent critic module. In this step, we select a subset comprising the top $p\%$ of trajectories based on their ratings. It's important to note that trajectories previously used for training in earlier iterations are excluded from selection. Additionally, we incorporate a set of chat data from the general domain to prevent overfitting. We maintain a 1:1 ratio between the trajectory data and the general data.

\textbf{Training Loss.}
We minimize the negative log-likelihood loss defined as follows:

\begin{equation}
\begin{split}
arg\min_{\theta}\sum_{n=1}^{N}\sum_{k=1}^{K}\sum_{m=1}^{M}-\text{log}\ \\
{P}_{\theta}(A_{m}^{(n,k)} | I^{(n,k)}A_{1}^{(n,k)}O_{1}^{(n,k)}\dots A_{m-1}^{(n,k)}O_{m-1}^{(n,k)})
\end{split}
\end{equation}

where $M$ is the maximum number of interaction rounds with the environment for each instruction. The $N$ is total number of instructions and $K$ is total number of trajectories to sample for each instruction. 

\subsection{Implementation Details}
For trajectory sampling, we sample 5 trials for each instruction, and each trial allows at most 5 rounds. For critic module top $10\%$ trajectories of highest scores are selected for training the agents. For model training, we set initial learning rate of $5e^{-5}$ and gradually reduces to $5e^{-6}$ using cosine annealing schedule with no warm-up steps when performing SFT. The Yi-34B \cite{young2024yi} model is used to implement the critic module.

\section{Experiment}

\subsection{Dataset}
We have selected the API-bank \cite{li2023api} as our benchmark dataset due to its extensive coverage of over 1,000 domains and more than 2,000 APIs in total. These APIs encompass a wide range of applications, including web search, health, calculation, weather forecasting, and more. For the purposes of this study, we utilize all 315 questions from the original dataset with released API implementations. Of these, 220 are employed for the agent evolution learning process, while the remaining 95 questions are set aside for performance evaluation. We present the accuracy based on the evaluation of these 95 questions. The accuracy is determined by the correctness of the answers, which are reviewed and verified by human evaluators.

\subsection{Ablation Study}
In this part, we evaluate effectiveness of individual components of the system including the trajectory sampling, critic module and performance of evolution. The actor employs the Yi-6B\cite{young2024yi} model for this study.

%\textbf{Scale of Trajectory Sampling.} The trajectory sampling procedure empowers the agents to explore their environment. We modulate the extent of this exploration using a factor $K$, as outlined in Equation~1. We experiment by varying the value of $K$ from 1 to 5 to observe how changes in the scale of environment exploration impact overall performance. The results of this experiment are presented in Table~[res]. Upon analyzing these results, we find that a larger $K$ value, which corresponds to a broader range of environment exploration, tends to yield higher accuracy, which confirms that environment exploration is a crucial step for agent evolution.

\textbf{Performance of Critic Module.} The primary role of the critic module is to identify a subset of trajectories that can potentially enhance the future performance of the agents. We evaluate the accuracy of this module by randomly selecting a number of samples rated by the critic and subjecting them to human evaluation. This allows us to assess the consistency between the critic's ratings and those of the human evaluator. The results of this evaluation are presented in Table~\ref{table:critic}. Our critic module achieves a precision of $70.0\%$ and a recall of $97.2\%$. These results suggest that the critic is effective in retaining most of the beneficial trajectories. However, the lower precision underscores the necessity of our iterative learning design, where only a minimal number of samples with the highest confidence scores are utilized for training.

\begin{table}[ht]
\centering
\begin{tabular}{|p{0.1\textwidth}|c|c|}
\hline
 & \textbf{Success (Critic)} & \textbf{Fail (Critic)} \\
\hline
\textbf{Success} \textbf{(Human)} & 35 & 1 \\
\hline
\textbf{Fail}  \textbf{(Human)} & 15 & 49  \\
\hline
\end{tabular}
\caption{Confusion matrix for critic module evaluation.}
\label{table:critic}
\end{table}

\begin{figure}[t]
  \includegraphics[width=\linewidth]{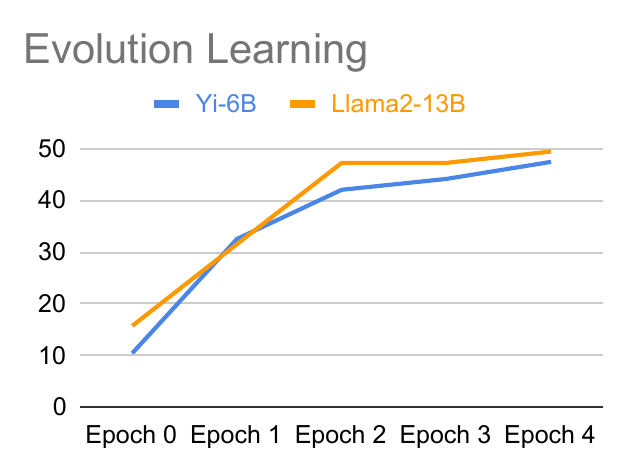}
  \caption{Accuracy results under different number of training iterations.}
  \label{fig:evolve}
\end{figure}

\textbf{Performance of Evolution.} Our agents evolve progressively over training iterations. To investigate the performance gain over iterations, we plot the accuracy w.r.t. different number of iterations for different base models with different model sizes, as shown in Figure~\ref{fig:evolve}. These results clearly confirms a steady improvement in accuracy over iterations, for both Yi-6B \cite{young2024yi} and Llama2-13B\cite{touvron2023llama2}.

\subsection{Benchmark}
In this section, we conduct experiments to compare with commercial GPT-4 model and the following open-source methods in the literature:

\textbf{ReAct-6B.} We implement the ReAct\cite{yao2022react} method with the same Yi-6B as base LLM. The detailed prompt is included in Appendix.

\textbf{API-Bank-7B}. Checkpoint model released by the original research paper \cite{li2023api} is used for evaluation.

\textbf{Yi-6B, Yi-34B}. Checkpoint models released by \cite{young2024yi}.

\textbf{Llama2-13B}. Checkpoint models released by \cite{touvron2023llama2}.

The comparison results are shown in Table~\ref{table:bench}. Firstly, our method has significantly outperformed the open-source models, without any expert-crafted training trajectories. Secondly, while at training stage our method relies on Yi-34B for critic module, our trained models of much smaller sizes (e.g. 6B and 13B) outperform the Yi-34B with a clear margin. Finally, with our evolution learning framework, our performance is close to that of strong commercial model GPT-4, suggesting that our method can effectively adapt to tasks.

\begin{table}[ht]
\centering
\begin{tabular}{|c|c|c|}
\hline
 \textbf{Model} & \textbf{ACC (\%)} \\
\hline
 GPT-4 & 51.6 \\
 \hline
ReAct-6B\cite{yao2022react} & 18.9 \\
API-Bank-7B \cite{li2023api} & 11.6 \\
Yi-6B \cite{young2024yi} & 10.5 \\
Yi-34B \cite{young2024yi} & 43.2 \\
Llama2-13B \cite{touvron2023llama} & 15.8 \\
\hline
\textbf{Ours-Yi-6B} & 47.5 \\
\textbf{Ours-Llama2-13B} & 49.5 \\
\hline
\end{tabular}
\caption{Benchmark Accuracy.}
\label{table:bench}
\end{table}

\section{Conclusion}
In conclusion, this paper presents a novel framework that enables the iterative self-evolution of agents, reducing dependence on expert-crafted trajectories or decisive environmental feedback. Our approach leverages a critic Large Language Model (LLM) to provide weak feedback, allowing agents to evolve progressively and learn incrementally. Despite potential errors from weak feedback, our iterative training process, which selects only a small number of high-confidence trials, ensures comprehensive environment exploration and effective learning. Notably, our framework achieves performance comparable to GPT-4 on the API-Bank public benchmark dataset using significantly smaller LLMs, demonstrating the efficacy of our approach.

\section{Limitations}
Our agents undergo iterative training, where each cycle requires multiple trials for environmental exploration. This method can be computationally demanding, especially for highly complex problems. Furthermore, the critic module's precision in evaluating trajectories remains somewhat limited, posing a potential constraint for applications that require exceptional accuracy.

\section{Ethical Statement}
This paper introduces a new training paradigm for agents, accompanied by ethical considerations. All data utilized in this study is publicly available, ensuring transparency and respect for privacy. Human evaluators were informed beforehand that their feedback might be used for product development and potentially published in a research paper. We are committed to acknowledging any limitations or potential biases in our findings. We affirm that no aspect of this research involved harm to individuals or misuse of personal data.

% \section{References}

\bibliography{acl_latex}

\appendix

\section{appendix}

\label{sec:appendix}

\end{document}